\documentclass[10pt,twocolumn,letterpaper]{article}

\usepackage[pagenumbers]{cvpr} 

\definecolor{cvprblue}{rgb}{0.21,0.49,0.74}
\usepackage[pagebackref,breaklinks,colorlinks,allcolors=blue]{hyperref}

\usepackage[vlined,linesnumbered,ruled]{algorithm2e}
\usepackage{multirow}
\usepackage{array}
\newcolumntype{C}[1]{>{\centering\arraybackslash}p{#1}}
\def\paperID{*****} 
\def\confName{CVPR}
\def\confYear{2025}

\title{The 1st Solution for 7th LSVOS RVOS Track: SaSaSa2VA}

\author{Quanzhu Niu$^{1*}$\quad
    Dengxian Gong$^{1*}$\quad
    Shihao Chen$^{1*}$\quad
    Tao Zhang$^{1*}$\\
    ~Yikang Zhou$^{1}$\quad
    Haobo Yuan$^{2}$\quad
    Lu Qi$^{1}$\quad
    Xiangtai Li$^{3}$\quad   
    Shunping Ji$^{1\dag}$ \vspace{3mm}\\
    {$^{1}$Wuhan University\quad
    $^{2}$University of California, Merced\quad
    $^{3}$Nanyang Technological University}\\
}

\begin{document}
\maketitle
\begin{abstract}
\renewcommand{\thefootnote}{}
\footnote{$^{*}$Equal contribution.}
\footnote{$^{\dag}$Corresponding author.}
Referring video object segmentation (RVOS) requires segmenting and tracking objects in videos conditioned on natural-language expressions, demanding fine-grained understanding of both appearance and motion. Building on Sa2VA, which couples a Multi-modal Large Language Model (MLLM) with the video segmentation model SAM2, we identify two key bottlenecks that limit segmentation performance: sparse frame sampling and reliance on a single~\texttt{[SEG]} token for an entire video. We propose \textbf{S}egmentation \textbf{A}ugmented and \textbf{S}elective \textbf{A}veraged \textbf{Sa2VA} (\textbf{SaSaSa2VA}) to address these issues. On the 7th LSVOS Challenge (RVOS track), SaSaSa2VA achieves a $\mathcal{J\&F}$ of \textbf{67.45}, ranking \textbf{first} and surpassing the runner-up by 2.80 points. This result and ablation studies demonstrate that efficient segmentation augmentation and test-time ensembling substantially enhance grounded MLLMs for RVOS. The code is released in Sa2VA repository:~\url{https://github.com/bytedance/Sa2VA}.
\end{abstract}    
\section{Introduction}
\label{sec:intro}

ICCV 2025 Large-scale Video Object Segmentation (LSVOS) Challenge has three tracks: Complex VOS on MOSEv2~\cite{MOSEv2}, targeting realistic, cluttered scenes with small, occluded, reappearing, and camouflaged objects under adverse conditions; VOS on MOSE~\cite{MOSE}, focusing on challenging, long and diverse videos; and RVOS on MeViS~\cite{MeViS,ding2025mevis}, assessing referring video object segmentation. 

Referring video object segmentation (RVOS) aims to segment and track objects in a video conditioned on a natural-language expression, and remains highly challenging. MeViS~\cite{MeViS,ding2025mevis} is a benchmark for RVOS that emphasizes motion-centric linguistic descriptions, making it more demanding than datasets~\cite{khoreva2019video, seo2020urvos} primarily driven by appearance-based expressions. Solving motion-expression–driven RVOS requires fine-grained video understanding together with strong video segmentation capabilities.

Recently, Multi-modal Large Language Models (MLLMs)~\cite{Qwen-VL, Qwen2-VL, Qwen2.5-VL, chen2024internvl, chen2024far, chen2024expanding, zhou2025they, internvl3, internvl3_5} have shown remarkable progress in image and video comprehension, including holistic scene understanding, recognition of object attributes and actions, and reasoning over inter-object relations.
In parallel, the video segmentation foundation model SAM2~\cite{ravi2024sam} substantially surpasses prior approaches~\cite{zhang2023dvis, zhang2025dvis++, zhou2024dvis, li2023tube, xu2024rap, zhang20231st_pvuw, zhang20231st_lsvos, niu2025} in both accuracy and generalization, enabled by a powerful data engine.
Grounded MLLMs~\cite{lai2024lisa, zhang2024omg, yan2024visa} further demonstrate that instruction-following segmentation can be achieved by coupling MLLMs with expert segmenters~\cite{kirillov2023segment, li2024omg, yuan2024mamba}.
Sa2VA~\cite{yuan2025sa2va} integrates the state-of-the-art MLLM InternVL 2.5~\cite{chen2024expanding} with SAM2~\cite{ravi2024sam}, yielding strong performance in image/video understanding and segmentation.
\begin{table}[t!]
    \centering
    \begin{tabular}{r|l|cc|c}
    \toprule
    Rank & Team / Username &  $\mathcal J$ & $\mathcal F$&  $\mathcal{J\&F}$\\
    \midrule
    \rowcolor{lightgray} \textbf{\# 1} & \textbf{SaSaSa2VA}   & \textbf{63.95} & \textbf{70.95}&\textbf{67.45}\\
    \# 2 & Transsion  & 61.29 & 68.01& 64.65\\
    \# 3 & dytino  & 61.06 & 67.22& 64.14\\
    \# 4 & heshuai  & 58.99 & 65.44& 62.22\\
    \# 5 & DanielLi  & 56.67 & 62.99& 59.83\\
    \bottomrule
    \end{tabular}
    \caption{\textbf{Leaderboard of the 7th LSVOS Challenge (RVOS track) in ICCV 2025.} Our \textbf{SaSaSa2VA} team achieves a $\mathcal{J\&F}$ score of \textbf{67.45} and wins first place.}
    \label{tab:leaderboard}
\end{table}

However, Sa2VA’s~\cite{yuan2025sa2va} segmentation potential is not fully realized. Sparse frame sampling limits the MLLM’s ability to model global spatiotemporal context, and relying on a single~\texttt{[SEG]} token to represent the entire video hampers robustness to temporal variations in object position, shape, and even appearance or disappearance.

\begin{figure*}[t]
  \centering
  \includegraphics[width=0.97\linewidth]{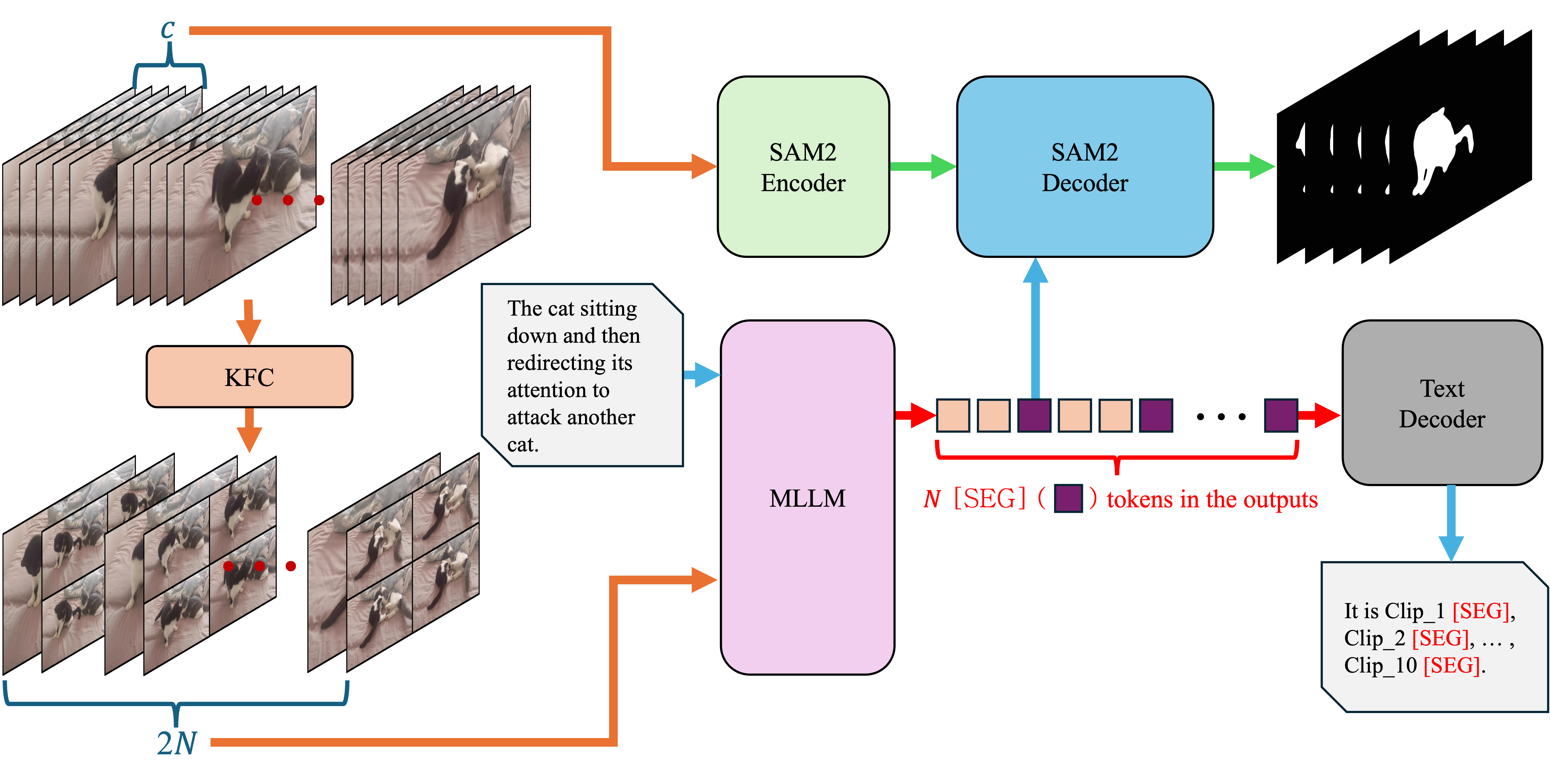}
  \caption{\textbf{Overview of Segmentation Augmentation in SaSaSa2VA. }We design Key Frame Compression (KFC) strategy and Scaling~\texttt{[SEG]} tokens strategy over Sa2VA~\cite{yuan2025sa2va}. A $T$-frame video is divided into $N$ non-overlapping clips, each containing $c = g^2{+}1$ frames. The frames passed to the MLLM are then compressed via KFC. The MLLM outputs $N$~\texttt{[SEG]} tokens, each corresponding to one clip. For a given clip, conditioned on the original $c$ frames and the hidden state of its~\texttt{[SEG]} token, SAM2 decodes the masks for that clip. In this figure, $c$ is set to $5$, resulting $g=2$.  }
  \label{fig:sa}
\end{figure*}

To address these issues, we introduce \textbf{S}egmentation \textbf{A}ugmented and \textbf{S}elective \textbf{A}veraged \textbf{Sa2VA} (\textbf{SaSaSa2VA}) in this challenge. Our Segmentation Augmentation improves the trade-off between efficiency and global video understanding; in particular, we employ Key Frame Compression and increase the number of~\texttt{[SEG]} tokens to better capture diverse temporal dynamics. At inference time, we apply test-time augmentation with five complementary sampling strategies and a Selective Averaging ensemble to exploit the strengths of different predictions.

With these designs, SaSaSa2VA achieves strong RVOS performance. As shown in~\cref{tab:leaderboard}, it attains a $\mathcal{J\&F}$ score of \textbf{67.45}.
Finally, our approach secures first place in the competition, underscoring both the promise of grounded MLLMs and the effectiveness of the proposed augmentation strategies.

\section{SaSaSa2VA}
\label{sec:method}


\subsection{Baseline: Sa2VA}
\label{sec:method_sa2va}
\noindent\textbf{Meta Architecture. }Sa2VA~\cite{yuan2025sa2va} comprises an Multi-modal Large Language Model (MLLM)~\cite{chen2024expanding} and SAM2~\cite{ravi2024sam}. The MLLM accepts images, videos, and text instructions as input and produces text responses conditioned on the instructions. When the user instruction requests segmentation results, the text response includes the segmentation token~\texttt{[SEG]}. The hidden state of the segmentation token serves as an implicit prompt, which SAM2 converts into object segmentation masks at both the image and video levels.

\noindent\textbf{MLLM. }Sa2VA adopts InternVL 2.5~\cite{chen2024expanding}. InternVL 2.5 follows a LLaVA-like~\cite{liu2023visual} architecture composed of an InternVIT~\cite{chen2024internvl}, an MLP projector, and a Large Language Model (LLM)~\cite{cai2024internlm2technicalreport, qwen2025qwen25technicalreport}. Images and videos are encoded by InternViT~\cite{chen2024internvl} into visual tokens, which are projected by an MLP and combined with text tokens as input to the LLM. The LLM autoregressively generates text responses that may include~\texttt{[SEG]} tokens. The hidden state of the~\texttt{[SEG]} token from the last LLM transformer layer is processed by an MLP to form the prompt input to SAM2~\cite{ravi2024sam}.

\noindent\textbf{SAM2. }SAM2~\cite{ravi2024sam} produces object segmentation masks for selected high-resolution video frames based on the segmentation prompts from the MLLM. It then propagates these frame-level masks to obtain object segmentation results for the entire video.

\subsection{Segmentation Augmentation}
\label{sec:method_sa}

\noindent\textbf{Limitations of Sa2VA. }To reduce time and memory consumption, only five frames are sampled per video during Sa2VA~\cite{yuan2025sa2va} training, and each video uses a single~\texttt{[SEG]} token to transmit information between the MLLM and SAM2. During inference, the MLLM processes only five frames for video understanding, and a single~\texttt{[SEG]} token is then used to propagate masks across the entire video. Sampling only five frames inevitably limits the MLLM's ability to capture global video context, and relying on a single~\texttt{[SEG]} token to convey segmentation information for the whole video struggles to accommodate temporal changes in object position, shape, and even appearance/disappearance. Consequently, this design imposes limitations on segmentation performance.

\noindent\textbf{Overview. }As illustrated in~\cref{fig:sa}, we design several Segmentation Augmentation strategies over Sa2VA~\cite{yuan2025sa2va}, including Key Frame Compression (KFC) and Scaling~\texttt{[SEG]} tokens. Details are described below.

\noindent\textbf{Key Frame Compression. }To balance spatiotemporal efficiency with the MLLM's global understanding of videos, we propose a Key Frame Compression (KFC) scheme. We sample $T = N\times c$ frames from the original video and denote the sequence as $V=\{I_1, I_2,\dots, I_{T}\}$, where each frame $I_t\in \mathbb{R}^{H\times W \times 3}$ is an RGB image at time step $t$. We then divide the sequence into $N$ non-overlapping clips, each containing $c = g^2+1$ frames. Each clip is denoted as $C^{i} = \{I^{i}_1, I^{i}_2,\dots, I^{i}_{c} \}$ for $i= 1, 2, \dots, N$. In each clip $C^{i}$, the first frame $I^{i}_1$ is the key frame, and the remaining $c-1=g^2$ frames $\{I^{i}_2, I^{i}_3,\dots, I^{i}_{c} \}$ are compressed. Specifically, we tile these $g^2$ frames into a $g\times g$ grid image $I^i_{cat}\in\mathbb{R}^{gH\times gW \times 3}$ in row-major order (left to right, top to bottom), and resize the grid back to $H\times W$ to obtain the compressed image $I^i_{com}\in\mathbb{R}^{H\times W \times 3}$:
\begin{equation}
    I^i_{cat} = \mathrm{concatenate}(I^{i}_2, I^{i}_2,\dots, I^{i}_{c}),
\end{equation}
\begin{equation}
    I^i_{com} = \mathrm{resize}(I^i_{cat}).
\end{equation}

In this way, each clip $C^{i}$ sends only one key frame and one compressed image $\{I^{i}_1, I^i_{com}\}$ to the MLLM, reducing a video with $T$ frames to just $2N$ images. This approach preserves global video information while mitigating redundant attention to adjacent frames.
                        
\begin{table*}[h!]
\centering
\begin{tabular}{p{1.1cm}p{2.5cm}||C{1.2cm}C{1.2cm}C{1.2cm}C{1.2cm}C{1.2cm}C{1.2cm}C{1.2cm}}
\toprule
\multicolumn{2}{l||}{\multirow{2}{*}{}}                        & \multicolumn{7}{c}{weights}                                                            \\ \specialrule{0em}{1pt}{1pt} \cline{3-9} \specialrule{0em}{1pt}{1pt}
\multicolumn{2}{l||}{}                                         & \multicolumn{2}{c|}{w/o SA} & \multicolumn{5}{c}{Selective Averaging} \\ \midrule \midrule
\multirow{5}{*}{14B}                           & Uniform      &            & \multicolumn{1}{c|}{}           & 2       & 2       & 2       & 1  & 1    \\
                                               & Uniform+     &            & \multicolumn{1}{c|}{}           & 2.5     & 2.5     & 2.5     & 1  & 1    \\
                                               & Q-frame      &            & \multicolumn{1}{c|}{}           &         &         &         & 1  & 1    \\
                                               & Wrap-around  &            & \multicolumn{1}{c|}{}           &         &         &         & 1  & 1    \\
                                               & Wrap-around+ &            & \multicolumn{1}{c|}{}           &         &         &         & 1  & 1    \\ \midrule
\multirow{5}{*}{26B}                           & Uniform      &            & \multicolumn{1}{c|}{}           & 1       & 1       & 1       & 1  & 1.5  \\
                                               & Uniform+     & 1          & \multicolumn{1}{c|}{}           & 1       & 1       & 1       & 1  & 1.5  \\
                                               & Q-frame      &            & \multicolumn{1}{c|}{}           & 1       & 1       & 1       & 1  & 1.5  \\
                                               & Wrap-around  &            & \multicolumn{1}{c|}{}           & 1       & 1       & 1       & 1  & 1.5  \\
                                               & Wrap-around+ &            & \multicolumn{1}{c|}{}           & 1       & 1       & 1       & 1  & 1.5  \\ \midrule
\multicolumn{1}{l}{\multirow{5}{*}{26B\ddag}} & Uniform      &            & \multicolumn{1}{c|}{}           &         &         & 1       & 1  & 1.2  \\
\multicolumn{1}{l}{}                           & Uniform+     &            & \multicolumn{1}{c|}{1}          &         & 2       & 1       & 1  & 1.2  \\
\multicolumn{1}{l}{}                           & Q-frame      &            & \multicolumn{1}{c|}{}           &         &         &         & 1  & 1.2  \\
\multicolumn{1}{l}{}                           & Wrap-around  &            & \multicolumn{1}{c|}{}           &         & 2       & 1       & 1  & 1.5  \\
\multicolumn{1}{l}{}                           & Wrap-around+ &            & \multicolumn{1}{c|}{}           &         &         & 1       & 1  & 1.5  \\ \midrule \midrule
\multicolumn{2}{c||}{$\mathcal J$}                                        & 62.00  & \multicolumn{1}{c|}{62.18}           &   63.21 &  \textbf{63.95}  & 63.69  & 63.82   &  63.77    \\
\multicolumn{2}{c||}{$\mathcal F$}                                        & 68.91   & \multicolumn{1}{c|}{69.15}           &   70.39& \textbf{70.95}  & 70.66 & 70.84   &  70.71    \\ \midrule
\multicolumn{2}{c||}{$\mathcal{J\&F}$}                                     & 65.46      & \multicolumn{1}{c|}{65.66}      & 66.80   & \textbf{67.45}   & 67.18   & 67.33   & 67.24   \\
\bottomrule
\end{tabular}
\caption{\textbf{Ablation on weighting schemes for Selective Averaging.} The scores are on the challenge split. ``\ddag'' indicates training without referring image segmentation datasets, and ``w/o SA'' denotes inference without Selective Averaging. }
\label{tab:ablation_study}
\end{table*}

\begin{table}[]
\centering
\begin{tabular}{l|l|cc}
\toprule
\multicolumn{2}{l|}{}                        & \multicolumn{1}{c}{valid\_u} & \multicolumn{1}{c}{valid} \\ \midrule \midrule
\multicolumn{2}{l|}{w/o SA (Sa2VA~\cite{yuan2025sa2va})}          & \multicolumn{1}{c}{61.8}     & \multicolumn{1}{c}{52.1}  \\ \midrule
\multirow{5}{*}{Ours} & Uniform      & 70.95                        & 66.52                     \\
                    & Uniform+     & 71.28                        & \textbf{66.78}                     \\
                    & Q-frame      & 71.35                        & 66.31                     \\
                    & Wrap-around  & 71.26                        & 66.33                     \\
                    & Wrap-around+ & \textbf{71.37}                        & 66.21       \\              
\bottomrule
\end{tabular}
\caption{\textbf{Ablation on Segmentation Augmentation.} We report the $\mathcal{J\&F}$ scores of the 26B model on the MeViS~\cite{MeViS,ding2025mevis} valid\_u and valid splits. ``w/o SA'' denotes training without Segmentation Augmentation, i.e., the Sa2VA baseline~\cite{yuan2025sa2va}.}
\label{tab:ablation_study1}
\end{table}
\noindent\textbf{Scaling~\texttt{[SEG]} tokens. }To handle diverse temporal variations of the objects, we increase the number of~\texttt{[SEG]} tokens. Specifically, we assign one~\texttt{[SEG]} token to each clip $C^{i}$, so the MLLM produces $N$~\texttt{[SEG]} tokens per video. The hidden states of these tokens are denoted by $S=\{s^1, s^2, \dots , s^N\}, s^i \in \mathbb{R}^d$, where $d$ is the hidden dimension. In SAM2, $s^i$ is used to decode the masks ($M^i=\{m^i_1, m^i_2, \dots, m^i_{c}\}, m^i_j \in \{0,1\}^{H\times W}$) of the object within clip $C^i$. The process is described by:
\begin{equation}
    S = \mathrm{MLLM}(I^1_1, I^1_{com}, I^2_1, I^2_{com}, \dots, I^N_1, I^N_{com}),
\end{equation}
\begin{equation}
    m^i_j = \mathrm{SAM2}(I^i_j, s^i).
\end{equation}

Specifically, during training, we supervise only the mask $m^i_1$ of the key frame $I^i_1$ in each clip.

\subsection{Test-time Augmentation}

\noindent\textbf{Inference sampling strategies. }During training, we sample $T = N\times c$ frames per video using a specific procedure, whereas at inference we must accommodate videos of varying lengths. To this end, we design five sampling strategies, each exhibiting advantages for different videos.

\begin{itemize}
    \item \textbf{Uniform. }Regardless of video length, the video is evenly divided into $N$ ori-clips. In each ori-clip, uniformly sample $c$ frames as the clip sent to the MLLM. In SAM2, the frames corresponding to each ori-clip need to be decoded using the associated $s$.
    \item \textbf{Uniform+. }Building on the Uniform strategy, for videos whose original length is shorter than $T$ frames, some frames near clip boundaries have 2 corresponding~\texttt{[SEG]} tokens. We average the masks from the 2 results.
    \item \textbf{Q-frame. }We use the method in~\cite{qframe} to select the top $T$ frames most related to the text prompt. The selected frames are then sorted in temporal order and processed with Uniform+.
    \item \textbf{Wrap-around. }Given a target of $T$ frames, sample cyclically using indices $i\bmod T_{ori}$, where $T_{ori}$ is the original video length. If $T_{ori}\geq T$, this yields the first $T$ frames; the masks beyond $T$ are propagated by SAM2's memory. If $T_{ori}< T$, it wraps around and repeats until $T$ frames are collected, and the selected frames are then sorted in temporal order.
    \item \textbf{Wrap-around+. }When $T_{ori}< T$, we use Wrap-around strategy. When $T_{ori}\geq T$, we instead use Uniform strategy.
\end{itemize}

\noindent\textbf{Selective Averaging. }Different inference sampling strategies perform differently across videos, and models of different scales also differ in their video understanding, leading to variations in final scores~\cite{fang20251stsolution4thpvuw}. To leverage their complementary strengths, we adopt a Selective Averaging scheme. For each mask, the results from different models and sampling methods are weighted and averaged. If the weighted average value of a pixel exceeds 0.5, its mask value is set to 1; otherwise, it is set to 0.
\section{Experiments}

\subsection{Implementation Details}

The baseline models we use are Sa2VA-14B and Sa2VA-26B~\cite{yuan2025sa2va}. The MLLM of Sa2VA-14B is InternVL 3.5-14B~\cite{internvl3_5}, and the MLLM of Sa2VA-26B is InternVL 2.5-26B~\cite{chen2024expanding}. We finetune Sa2VA using the Segmentation Augmentation strategies described in~\cref{sec:method_sa}. We fix $T=100$ and $N=10$, resulting in $c=10$ and $g=3$. During finetuning, we use referring image segmentation datasets including RefCOCO~\cite{refcoco}, RefCOCO+~\cite{refcoco}, and RefCOCOg~\cite{refcocog}, as well as referring video object segmentation datasets including MeViS~\cite{MeViS,ding2025mevis}, Ref-YTVOS~\cite{seo2020urvos}, ReVOS~\cite{yan2024visa}, and Ref-SAV~\cite{yuan2025sa2va}. Details of Selective Averaging are provided in~\cref{sec:ablation_study}. For the challenge, we adopt the best-performing strategy.

\subsection{Main Results}\label{sec:main_result}

The final challenge results are shown in Table~\ref{tab:leaderboard}. Our method achieved the highest score in the challenge, substantially outperforming other teams across all metrics ($\mathcal{J\&F}$, $\mathcal J$, and $\mathcal F$), achieving a $\mathcal{J\&F}$ of \textbf{67.45} and surpassing the second place by \textbf{2.80}. This demonstrates the superiority of our approach.

\subsection{Ablation Study}
\label{sec:ablation_study}

\textbf{Segmentation Augmentation.} As shown in~\cref{tab:ablation_study1}, applying our Segmentation Augmentation consistently improves performance across all inference sampling strategies, yielding gains of more than \textbf{10 }$\mathcal{J\&F}$ points over the baseline~\cite{yuan2025sa2va}. This demonstrates that Segmentation Augmentation is the most effective strategy in this challenge.

\noindent\textbf{Weighting schemes for Selective Averaging.} We compare the weighting schemes of various Selective Averaging strategies and report the challenge set scores in~\cref{tab:ablation_study}. Results with Selective Averaging are significantly better than without, improving $\mathcal{J\&F}$ by about \textbf{2} points. This verifies that ensembling across different inference strategies and model scales via voting yields substantial gains, demonstrating the effectiveness of the Selective Averaging method.

\section{Conclusion}
\label{sec:conclusion}

In this report, we enhance grounded MLLMs for RVOS by addressing two practical bottlenecks — limited temporal coverage and under-expressive segmentation prompting. Our SaSaSa2VA introduces Segmentation Augmentation to improve global video understanding while remaining efficient, and employs Selective Averaging at inference to robustly fuse complementary predictions. The approach achieves state-of-the-art performance on the 7th LSVOS Challenge (RVOS track), validating both its accuracy and practicality. Beyond competitive results, our findings highlight the importance of aligning MLLM reasoning with video segmentation.

{
    \small
    \bibliographystyle{ieeenat_fullname}
    \bibliography{main}
}


\end{document}